\documentclass{article}

\usepackage[nonatbib,final]{nips_dlps_2017}

\usepackage[utf8]{inputenc} 
\usepackage[T1]{fontenc}    
\usepackage{hyperref}       
\usepackage{url}            
\usepackage{booktabs}       
\usepackage{amsfonts}       
\usepackage{nicefrac}       
\usepackage{microtype}      
\usepackage{cite}			
\usepackage{graphicx}		
\usepackage{textcomp}		
\title{PSIque: Next Sequence Prediction of Satellite Images using a Convolutional Sequence-to-Sequence Network}

\author{
  Seungkyun Hong $^{*,1,2}$ \qquad
  Seongchan Kim $^{2}$ \qquad
  Minsu Joh $^{1,2}$ \qquad 
  Sa-kwang Song \thanks{Corresponding Author} $^{,1,2}$   \\
  $^{1}$Korea University of Science and Technology \\
  $^{2}$Korea Institute of Science and Technology Information \\
  \texttt{\{xo,sckim,msjoh,esmallj\}@kisti.re.kr}
}

\begin{document}

\maketitle

\begin{abstract}
  Predicting unseen weather phenomena is an important issue for disaster management. In this paper, we suggest a model for a convolutional sequence-to-sequence autoencoder for predicting undiscovered weather situations from previous satellite images. We also propose a symmetric skip connection between encoder and decoder modules to produce more comprehensive image predictions. To examine our model performance, we conducted experiments for each suggested model to predict future satellite images from historical satellite images. A specific combination of skip connection and sequence-to-sequence autoencoder was able to generate closest prediction from the ground truth image.
   
\end{abstract}

\section{Introduction}
Destruction caused by extreme weather events, such as cyclones and wildfires, has garnered increasing attention in recent decades, and the need to predict such events has required more descriptive nowcasting technologies.
Currently, numerical weather prediction (NWP) models are performed to forecast future weather situations;
still, these simulations require intensive computation and long-term cumulative datasets.
Further, although NWP have advantages of accuracy and widespread simulation performance, additional effort is required to fit these models to a responsive nowcasting system.

Recent improvements in earth observation systems have allowed us to observe the changes in global weather on an hourly basis via remote sensing imageries.
Since massive earth observation data with long durations are now available, several works have surveyed the effectiveness of weather events considering prediction tasks.

However, occasionally, there is a possibility of data loss from remote sensory systems.
For example, a geostationary satellite may fail to produce broad observation data owing to several issues such as image merging, satellite repositioning on the designated orbit, special missions such as solar or lunar eclipse, and so on.
These operational failures can result in satellite stations facing critical data loss. 
However, there could be possible sudden emergency situations wherein weather events prediction without sufficient data sources or even on nonexistent moments may be required. Given this background, it is unclear how event detection systems will forecast unseen or missed situations.

In this paper, we suggest an approach for predicting a visually unobserved situation by generating a plausible image by referencing historical time-series visual clues. We found that specific information propagation leads to better, reasonable prediction results.
The remainder of this paper is organized as followed. In the next section, we first discuss the recent research milestones in future image prediction techniques. Next, we present our proposed model structure for future weather situation prediction, and we conclude with comparison results and the conclusions.

\section{Related works}
A prediction task of future weather conditions is a joint task comprising two separate meteorological processes: 
(1) understanding visual clues for weather transition from spatiotemporal weather observations, and
(2) image restoration from low-dimensional encoded information into high-dimensional visual situations through time.

Recently, several studies focused on weather events prediction using either observation or simulation data.
Shi et al.\cite{1506.04214:Shi2015ConvLSTM} and Kim et al.\cite{Kim2017DeepRainCI} predicted future precipitation from historical multichannel radar reflectivity images with a convolutional long short-term memory (ConvLSTM) network\cite{Hochreiter1997LONGMEMORY} based on recurrent neural network (RNN)\cite{Elman1990FindingTime} and convolutional neural networks (CNNs).
Racah et al.\cite{1612.02095:RacahSemiSup} suggested an 3D convolutional autoencoder (AE) model for extreme weather events detection using 27-years CAM-5 simulation model results.
Hong et al.\cite{1708.03417:GlobeNet} surveyed multiple CNNs for predicting the coordinates of a typhoon eye from a single satellite image.
Kim et al.\cite{Kim2017MassiveEvents} suggested a tropical cyclone detection system based on GCM reanalysis data using 5-layer CNNs.
These works depend on historical visual datasets for predicting weather events.

However, it is not confirmed if unseen weather events can be predicted for specific occasions without observation data such as an observation failure and future events prediction. For the issue of missing/lost data, several studies on deep neural memory networks have been conducted for predicting undiscovered data. \cite{Elman1990FindingTime,Hochreiter1997LONGMEMORY}
Based on the LSTM-RNN, Srivastava et al.\cite{1502.04681:srivastava2015unsup} surveyed an unsupervised video representation on a fully-connected LSTM (fcLSTM) network with flatten data.
Accordingly, Patraucean et al.\cite{1511.06309:Patraucean2015Spatio} used intermediate differentiable memory on a temporal video auto-encoder network, and trained extensive optical flow transition differences through time between encoding and decoding steps.
Chen et al.\cite{1609.01006:Chen3D} considered a 3D biomedical image as the spatial data continuum and adopted a bi-directional LSTM structure for understanding 3D contexts beyond correlated 2D slices.

With the advantages of a memory network, several research studies also discovered the effectiveness of structural separation into discrete encoder and decoder units for both sequenced input and output.
Cho et al.\cite{1406.1078:cho2014learning} suggested an RNN encoder--decoder architecture associated with two separated RNNs as an encoder and divaricated decoder parts.
Accordingly, Sutskever et al.\cite{1409.3215:sutskever2014sequence} employed the structure of \cite{1406.1078:cho2014learning} and an adopting LSTM cell\cite{Hochreiter1997LONGMEMORY} on the overall topology, entitled sequence-to-sequence (Seq2Seq).
Next, Chorowski et al.\cite{1506.07503:chorowski2015attention} introduced an external memory in Seq2Seq called attention memory, as well as which is similar to a differentiable memory, which was used in a convolutional attention-based Seq2Seq\cite{1710.04515}. For image restoration issues, Mao et al.\cite{1603.09056:SkipConx} introduced the residual connectivity of the results of convolutional processes into the deconvolutional step for maximizing plausible image reconstruction. Enriched with abundant visual clues from deep convolutional filters, symmetric skip connections (SkipConx) helped achieve high performance as a denoising application, and as a conventional image autoencoder.

In conclusion, few surveys have been conducted on image reconstruction for remote sensing imagery, particularly on a satellite image dataset.
Therefore, we studied the initial structure of learning weather changes through time, with minimal available parameters.
In other words, we studied how changes in weather can be represented as fundamental structures of a memory network.

\section{Methodology}

\subsection{Model Architecture}
In this section, we discuss the architecture for predicting next one-hour satellite images using historical observation datasets.
To predict undiscovered situations, we used a basic Seq2Seq architecture that consists of two RNNs using an encoder-decoder framework.
In the encoder section, existing weather situation data are plotted and encoded as reduced information through time.
Next, the decoder receives outputs and network states from the encoder, and uses this information to predict future situations one-by-one.

Both the encoder and the decoder use a five-layer convolution operation to extract rich features or restore encoded information during deconvolution operations.
As convolutional filter weights of the encoder and the decoder could be repeatedly utilized, the encoder and the decoder must be able to save and restore their parameters within a scope of Seq2Seq memory networks because the model is not restricted to a specific number of input and output sequences.
The suggested model can therefore be used for flexible input/output configurations. 
However, a fixed time gap between each data instance (e.g., 1 hour gap per input) is suggested to achieve better prediction results.

As the basic LSTM network uses dense 1D information, we flattened the 2D image data into 1D tensors to fit LSTM cells.
To achieve better feature extraction from the spatial data space, we introduce convolutional LSTM cells into both the encoder and the decoder.
Further, we implement an auxiliary link in the image regeneration step---a symmetric skip connection from the convolutional encoder into the deconvolutional decoder---for better image restoration.

\begin{figure}[t]
\vspace{-1.1cm}
\begin{center}
\includegraphics[height=3.5cm]{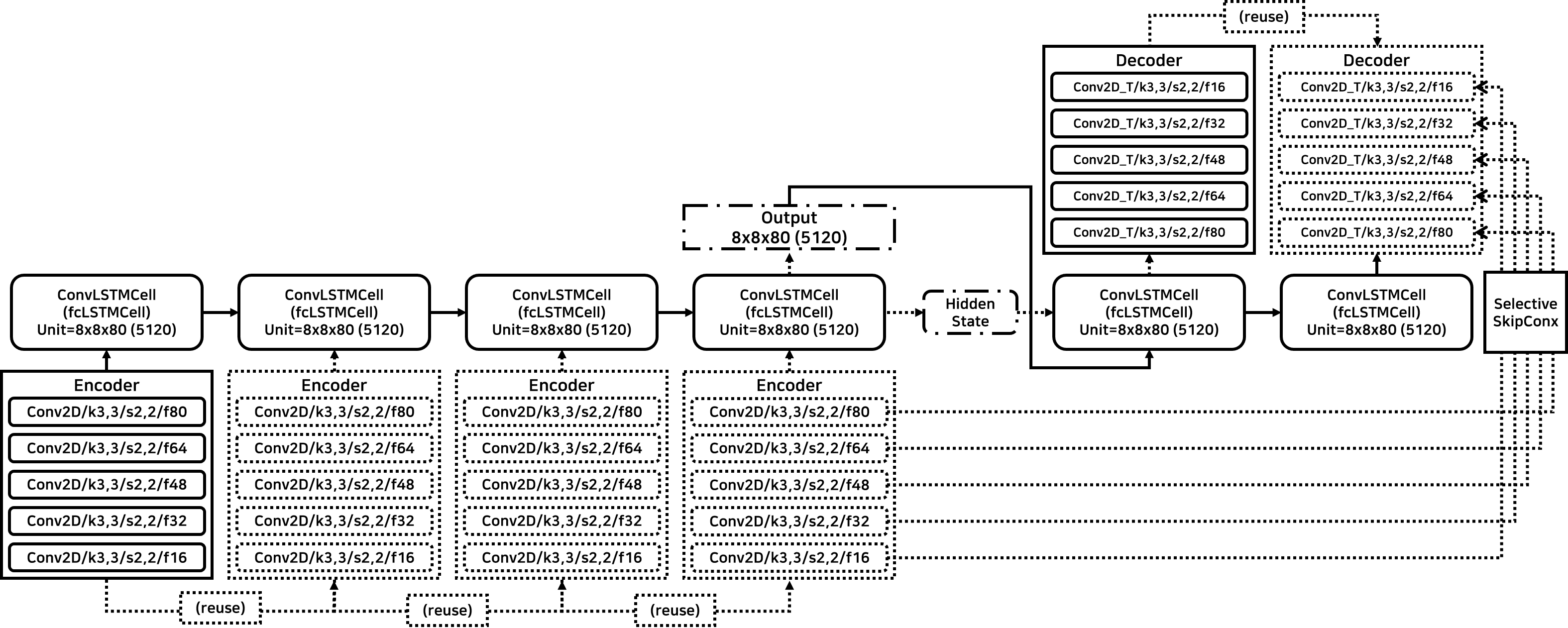}
\caption{Model topology for Seq2Seq Autoencoder with Selective Skip Connection (SkipConx)}
\end{center}
\vspace{-0.7cm}
\end{figure}

\subsection{Dataset}
We used four-channel infrared images observed from the COMS-1 satellite meteorological imager (MI) for weather situation prediction. This dataset contains data from April 2011 to June 2017, and it consists of only a partial subset owing to data collection issues. The pixel dimension of the image dataset was resized to $256 \times 256 \times 4$ for fast model training. Further, we designed a data sequencer gateway for the Seq2Seq autoencoder, which receives four-step information to interpret weather changes and produces two-step weather transitions. Meanwhile, a single instance of the data sequencer for training Seq2Seq autoencoder requires six iterations of satellite observation, i.e., one hour per image, for a total time of 6 h. Based on those sequence settings and the status of data collection, we constructed about 26,490 instances to train several Seq2Seq autoencoders and split the set with a ratio of 0.81:0.09:0.1 (for the training, validation, and test configurations).

\section{Evaluation}

\begin{figure}[h]
\vspace*{-0.6cm}
\begin{center}
\includegraphics[height=3.6cm]{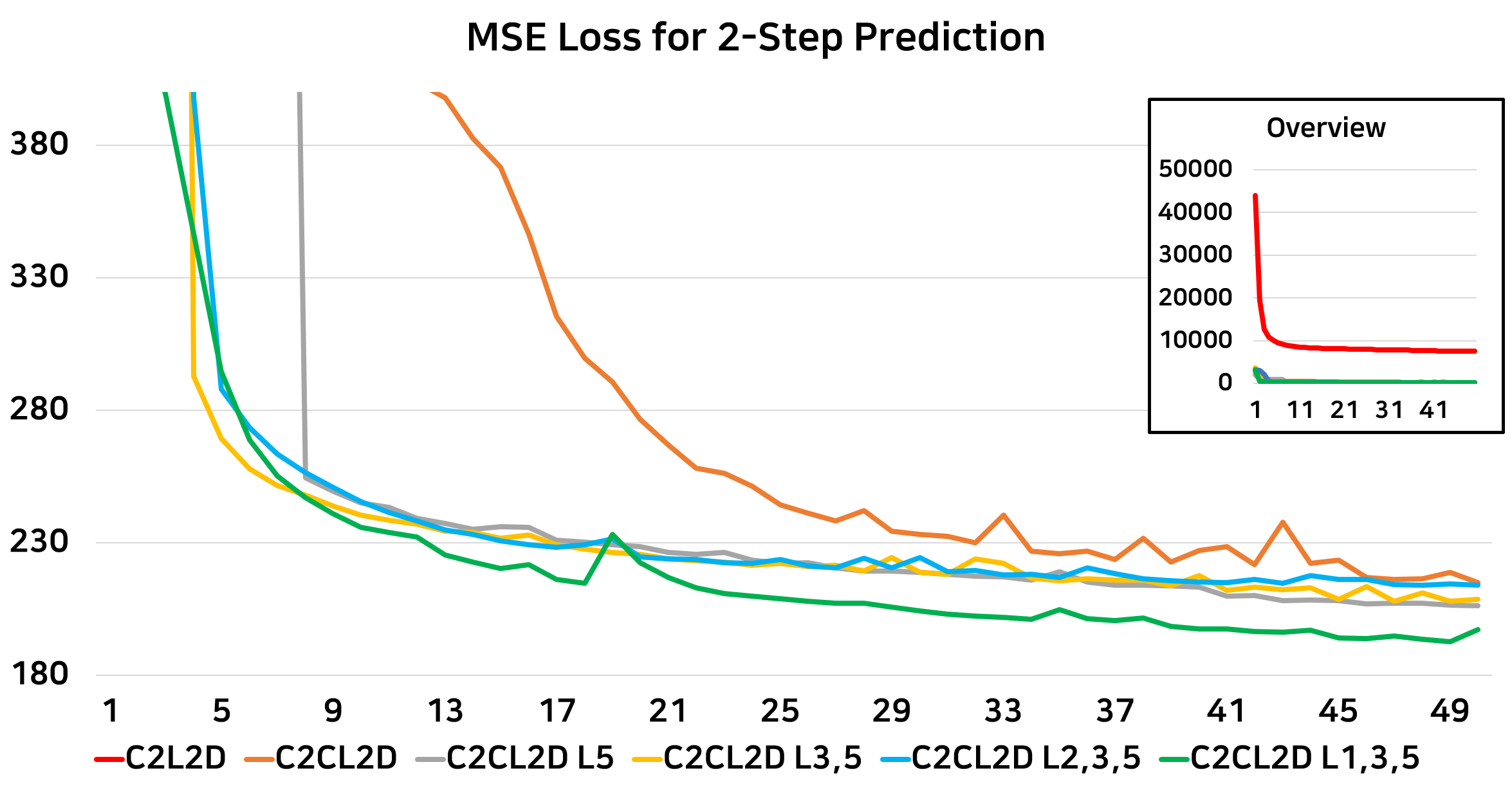}
\caption{Model loss of each model configuration. Lower scores show better prediction results.}
\end{center}
\label{fig:error}
\vspace*{-0.4cm}
\end{figure}

To evaluate the performance of the convolutional Seq2Seq model, we conducted multiple experiments to identify the best image prediction setting. A basic LSTM cell was used for image restoration (C2L2D), and a convolutional LSTM cell was also used (C2CL2D) for comparison. Next, a multiple skip connection was designed between the encoder and the decoder in the C2CL2D model (C2CL2D-SkipConx) to discover the optimal topology for plausible image restoration. Since our model uses five convolutional layers, we created several combinations of layers for selective symmetric skip connections: L5, L3/5, L2/3/5, and L1/3/5.

The accuracy of weather situation prediction models is defined as (\ref{exp:mse}), where \textit{P} is a set of pixels that consists of width × height and the longitude of the ground truth, and \textit{$\hat{P}$} is a set of pixels from the prediction result.
\begin{equation}
MSE_{Pixels} = \frac{1}{N} \sum_{n=1}^{N} {(P-\hat{P})}^2
\label{exp:mse}
\end{equation}

Each testing model was trained for 50 epochs and used an exponential linear unit (Elu) as an activation function for increasing nonlinearity. Further, the Adam optimizer \cite{1412.6980:Kingma2015Adam} is used for gradient optimization with an initial learning rate 1e-3 to deal with large pixel values. All our models use the toolkit TensorFlow \cite{1605.08695:TensorFlow} release 1.4 as the numerical computation framework with GPU acceleration support from the CUDA/cuDNN framework. For model training and inference acceleration, we used dual NVIDIA\textsuperscript{\textregistered} Tesla\textsuperscript{\textregistered} P100 GPU accelerators for boosting mini-batch training.

\begin{table}[h]
  \caption{MSE (Test) result of each model configuration after 50 epochs.}
  \label{rmse_full}
  \centering
  \begin{tabular}{lll}
    \toprule
    Model & Configuration & MSE \\
    \midrule
    C2L2D \ \ \ (vanilla) & fcLSTM & 7537.908 [baseline] \\
    C2CL2D (vanilla) & ConvLSTM & \ \ 215.039 \\
    C2CL2D (SkipConx) & ConvLSTM, L5 & \ \ 206.413 \\
    C2CL2D (SkipConx) & ConvLSTM, L3/5 & \ \ 208.698 \\
    C2CL2D (SkipConx) & ConvLSTM, L2/3/5 & \ \ 214.000 \\
    C2CL2D (SkipConx) & ConvLSTM, L1/3/5 & \ \ 197.222 \\
    \bottomrule
  \end{tabular}
\end{table}

\textbf{Fig. 2} and \textbf{Table. 1} shows the result of test errors for each situation prediction model. All test models converged at some point. Compared to the C2L2D model, other C2CL2D models almost converged around MSE 200. Unfortunately, the C2L2D model was trained too slowly and weakly, and the last test error among 50 epoch trainings was almost MSE 7538, which is almost 35 times larger than the other C2CL2D model configurations.

\begin{figure}[h]
\vspace*{-0.05cm}
\begin{center}
\includegraphics[height=4cm]{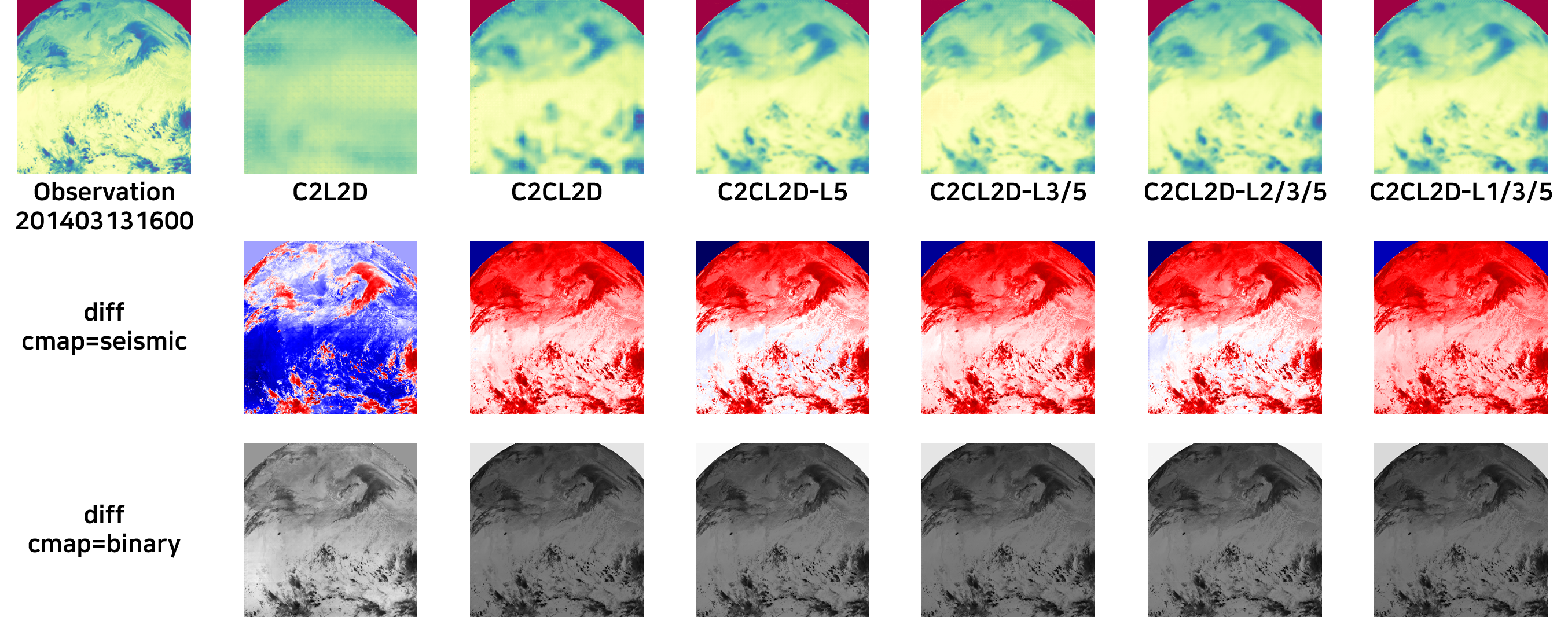}
\caption{Ground truth (actual observation) vs. Prediction through time on Seq2Seq}
\end{center}
\label{fig:decoderesult}
\vspace*{-0.4cm}
\end{figure}

\textbf{Fig. 3} shows an example of qualitative prediction results from all model configurations, compared with original observation data (201403131600 UTC, ch1-IR1).
The prediction results of the C2L2D model are almost unusable to figure out the weather situation, which show a lack of learning capability. In contrast, the C2CL2D model shows more comprehensive weather predictions than the C2L2D result. The C2CL2D model with the skip connection on Layer 1/3/5 achieved the lowest test error among all models, and it presented the most reliable weather prediction result.

\section{Conclusion}
In this work, we studied how a Seq2Seq-based convolutional autoencoder can predict unseen weather situations by referencing historical weather observation datasets. Further, we found that adding symmetric skip connection from the convolutional encoder to the deconvolutional decoder in the Seq2Seq autoencoder provides more reliable image prediction compared to bare LSTM or primitive convolutional LSTM connectivity.
Meanwhile, our proposed model still needs improvement in terms of image resolution for more comprehensive weather situation prediction. We have planned several optimizations, such as differentiable memory and usage of adversarial networks, to improve our work.

\section{Acknowledgement}
This work formed part of research projects carried out at the Korea Institute of Science and Technology Information (KISTI). (Project No. K-17-L03-C03: Construction of HPC-based Service Infrastructure Responding to National Scale Disaster; Project No. K-17-L05-C08: Research for Typhoon Track Prediction using an End-to-End Deep Learning Technique) We also gratefully acknowledge the support of NVIDIA Corporation with the donation of multiple GPUs used for this research.

\bibliographystyle{ieeetr}
\bibliography{nips2017-reference}

\end{document}